\documentclass[11pt]{article}
\usepackage{acl2016}
\usepackage{times}
\usepackage{latexsym}
\usepackage{graphicx} 
\usepackage{mathtools}
\usepackage{subfigure}
\usepackage{algorithm}
\usepackage{algorithmic}

\usepackage[procnames]{listings}
\usepackage{color}

\usepackage{booktabs} 

\aclfinalcopy 
\def\aclpaperid{253} 

\usepackage{hyperref}

\newcommand{\softmax}{\mathrm{softmax}}
\newcommand{\argmax}{\mathrm{argmax}}

\title{Latent Predictor Networks for Code Generation}
\author{
Wang Ling$\diamondsuit$ Edward Grefenstette$\diamondsuit$ Karl Moritz Hermann$\diamondsuit$
\\
{\bf Tom\'a\v{s} Ko\v{c}isk\'y$\diamondsuit$$\clubsuit$ Andrew Senior$\diamondsuit$ Fumin Wang$\diamondsuit$ Phil Blunsom$\diamondsuit$$\clubsuit$}
\\
\\
$\diamondsuit$Google DeepMind $\clubsuit$University of Oxford
\\
\{lingwang,etg,kmh,tkocisky,andrewsenior,awaw,pblunsom\}@google.com
}\date{}

\begin{document}

\maketitle

\begin{abstract}
Many language generation tasks require the production of text
conditioned on both structured and unstructured inputs. We
present a novel neural network architecture which generates an output sequence
conditioned on an arbitrary number of input functions. Crucially, our approach
allows both the choice of conditioning context and the granularity of
generation, for example characters or tokens, to be marginalised, thus
permitting scalable and effective training.  Using this framework, we address
the problem of generating programming code from a mixed natural language and
structured specification. We create two new data sets for this paradigm derived
from the collectible trading card games Magic the Gathering and Hearthstone. On
these, and a third preexisting corpus, we demonstrate that marginalising
multiple predictors allows our model to outperform strong benchmarks.

\end{abstract}

\section{Introduction}

The generation of both natural and formal languages often requires models conditioned on diverse predictors~\cite{Koehn:2007:MOS:1557769.1557821,Wong:2006:LSP:1220835.1220891}. Most models take the restrictive approach of employing a single predictor, such as a word softmax, to predict all tokens of the output sequence. To illustrate its limitation, suppose we wish to generate the answer to the question ``Who wrote The Foundation?'' as ``The Foundation was written by Isaac Asimov''. The generation of the words ``Issac Asimov" and ``The Foundation'' from a word softmax trained on annotated data is unlikely to succeed as these words are sparse. A robust model might, for example, employ one predictor to copy ``The Foundation'' from the input, and a another one to find the answer ``Issac Asimov'' by searching through a database. However, training multiple predictors is in itself a challenging task, as no annotation exists regarding the predictor used to generate each output token. Furthermore, predictors generate segments of different granularity, as database queries can generate multiple tokens while a word softmax generates a single token. In this work we introduce \emph{Latent Predictor Networks (LPNs)}, a novel neural architecture that fulfills these desiderata: at the core of the architecture is the exact computation of the marginal likelihood over latent predictors and generated segments allowing for scalable training.

\begin{figure}[t!]
\begin{center}
\centerline{\includegraphics[width=0.9\columnwidth,trim=0cm 0cm
    0cm 0cm]{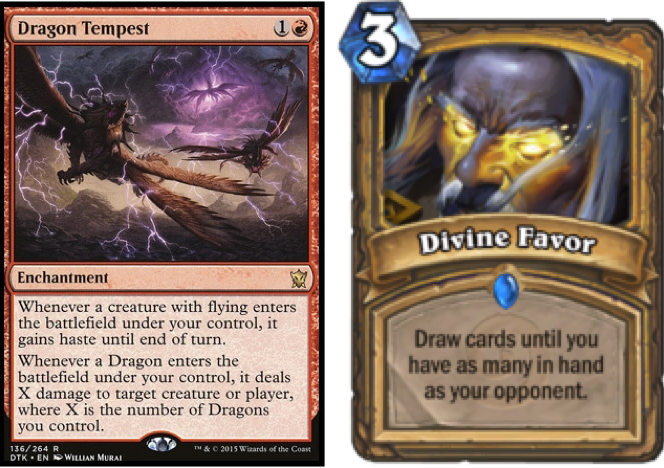}}
\caption{Example MTG and HS cards.}
\label{hs_img}
\end{center}
\end{figure}

We introduce a new corpus for the automatic generation
of code for cards in Trading Card Games~(TCGs), on which we validate our model~\footnote{Dataset available at https://deepmind.com/publications.html}. TCGs, such as Magic the
Gathering~(MTG) and Hearthstone~(HS), are games played between two players that
build decks from an ever expanding pool of cards. Examples of such cards are
shown in Figure~\ref{hs_img}.
Each card is identified by its attributes (e.g., name and cost) and
has an effect that is described in a text box. Digital implementations of these games implement the game logic, which includes the card effects. This is attractive from a data extraction perspective as not only are the data annotations naturally generated, but we can also view the card as a specification communicated from a designer to a software engineer.

This dataset presents additional challenges to prior work in code generation~\cite{Wong:2006:LSP:1220835.1220891,Jones:2012:SPB:2390524.2390593,lei-EtAl:2013:ACL2013,artzi-lee-zettlemoyer:2015:EMNLP,quirk:acl15}, including the handling of structured input---i.e.~cards are composed by multiple sequences (e.g., name and description)---and attributes (e.g., attack and cost), and the length of the generated sequences. Thus, we propose an extension to attention-based neural models~\cite{DBLP:journals/corr/BahdanauCB14} to attend over structured inputs. Finally, we propose a code compression method to reduce the size of the code without impacting the quality of the predictions.

Experiments performed on our new datasets, and a further pre-existing one, suggest that our extensions outperform strong benchmarks.

The paper is structured as follows: We first describe the data collection process~(Section \ref{sec:data}) and formally define our problem and our baseline method~(Section \ref{problem}). Then, we propose our extensions, namely, the structured attention mechanism~(Section \ref{sec:structure}) and the LPN architecture~(Section \ref{sec:lpn}). We follow with the description of our code compression algorithm~(Section \ref{sec:compression}). Our model is validated by comparing with multiple benchmarks~(Section \ref{sec:experiments}). Finally, we contextualize our findings with related work~(Section \ref{sec:rel}) and present the conclusions of this work~(Section \ref{sec:conclusion}).

\section{Dataset Extraction}
\label{sec:data}

We obtain data from open source implementations of two different TCGs, MTG in Java\footnote{\url{github.com/magefree/mage/}} and HS in Python.\footnote{\url{github.com/danielyule/hearthbreaker/}} The statistics of the corpora are illustrated in Table~\ref{tab:corpus}. In both corpora, each card is implemented in a separate class file, which we strip of imports and comments.
We categorize the content of each card into two different groups: \emph{singular fields} that contain only one value; and \emph{text fields}, which contain multiple words representing different units of meaning. In MTG, there are six singular fields (attack, defense, rarity, set, id, and health)
and four text fields (cost, type, name, and
description), whereas HS cards have eight singular
fields (attack, health, cost and durability, rarity,
type, race and class) and two text fields (name and description).
Text fields are tokenized by splitting on whitespace and punctuation,
with exceptions accounting for domain specific artifacts (e.g., Green mana is described as ``\{G\}'' in MTG).
Empty fields are replaced with a ``NIL'' token.

\begin{table}
\centering
\small
\begin{tabular}{@{}lrr@{}}
\toprule
{\bf } & {\bf MTG} & {\bf HS}\\
\midrule
Programming Language & Java & Python\\
\midrule
Cards & 13,297 & 665 \\
Cards (Train) & 11,969 & 533 \\
Cards (Validation) & 664 & 66 \\
Cards (Test) & 664 & 66 \\
\midrule
Singular Fields & 6 & 4 \\
Text Fields & 8 & 2 \\
\midrule
Words In Description (Average) & 21 & 7 \\
Characters In Code (Average) & 1,080 & 352 \\
\bottomrule
\end{tabular}
\caption{Statistics of the two TCG datasets.}\label{tab:corpus}
\end{table}

The code for the HS card in Figure~\ref{hs_img} is shown in Figure~\ref{df_code}. The effect of ``drawing cards until the player has as many cards as the
opponent'' is implemented by computing the difference between the players' hands
and invoking the draw method that number of times. This illustrates that
the mapping between the description and the code is non-linear, as no information is given in the text regarding the specifics of the implementation.

\begin{figure}[ht!]
\begin{center}
\begin{lstlisting}
class DivineFavor(SpellCard):
    def __init__(self):
        super().__init__("Divine Favor", 3,
    CHARACTER_CLASS.PALADIN, CARD_RARITY.RARE)

    def use(self, player, game):
        super().use(player, game)
        difference = len(game.other_player.hand)
       - len(player.hand)
        for i in range(0, difference):
            player.draw()
\end{lstlisting}
\end{center}
\caption{Code for the HS card ``Divine Favor''.}
\label{df_code}
\end{figure}

\section{Problem Definition}
\label{problem}
Given the description of a card $x$, our decoding problem is to find the code $\hat{y}$ so that:
\begin{equation}
\label{max}
\hat{y} = \underset{y}\argmax \log P(y \mid x)
\end{equation}
Here $\log P(y \mid x)$ is estimated by a given model.
We define $y=y_1..y_{|y|}$ as the sequence of characters of the code with
length $|y|$. We index each input field with $k={1..|x|}$, where $|x|$
quantifies the number of input fields. $|x_k|$ denotes the number of tokens in $x_k$ and $x_{ki}$ selects the $i$-th token.

\section{Structured Attention}
\label{sec:structure}

\paragraph{Background} When $|x|=1$, the attention model of~\newcite{DBLP:journals/corr/BahdanauCB14} applies.
Following the chain rule,
$\log P(y|x) = \sum_{t=1..|y|} \log P(y_t |  y_1..y_{t-1}, x)$,
each token $y_t$ is predicted conditioned on the previously generated
sequence $y_1..y_{t-1}$ and input sequence $x_1=x_{11}..x_{1|x_1|}$. Probability are estimated with a
softmax over the vocabulary $Y$:
\begin{equation}
\label{char_softmax}
p(y_t|y_1..y_{t-1}, x_1) = \underset{y_t \in Y}\softmax(\mathbf{h}_t)
\end{equation}
where $\mathbf{h}_t$ is the Recurrent Neural Network (RNN) state at
time stamp $t$, which is modeled as $g(\mathbf{y}_{t-1},\mathbf{h}_{t-1},\mathbf{z}_t)$.
$g(\cdot)$ is a recurrent update function for generating the new state
$\mathbf{h}_t$ based on the previous token
$\mathbf{y}_{t-1}$, the previous state $\mathbf{h}_{t-1}$, and the input text
representation $\mathbf{z}_{t}$. We implement $g$ using a Long
Short-Term Memory (LSTM) RNNs~\cite{Hochreiter:1997:LSM:1246443.1246450}.

The attention mechanism generates the representation of the input
sequence $\mathbf{x} = x_{11}..x_{1|x_1|}$, and $\mathbf{z}_{t}$ is computed as
the weighted sum $\mathbf{z}_{t} = \sum_{i = {1..|x_1|}} a_i h(x_{1i})$,
where $a_i$ is the attention coefficient obtained for token $x_{1i}$ and $h$ is a
function that maps each $x_{1i}$ to a continuous vector. In general, $h$ is a
function that projects $x_{1i}$ by learning a lookup table, and then embedding
contextual words by defining an RNN. Coefficients $a_i$ are computed
with a softmax over input tokens $x_{11}..x_{1|x_1|}$:
\begin{equation}
\label{alignment}
a_i = \underset{x_{1i} \in x}\softmax\left(v(h(x_{1i}),\mathbf{h}_{t-1})\right)
\end{equation}
Function $v$ computes the affinity of each token $x_{1i}$ and the current
output context $\mathbf{h}_{t-1}$. A common implementation of $v$ is to apply a
linear projection from $h(x_{1i}):\mathbf{h}_{t-1}$ (where $:$ is the concatenation operation)
into a fixed size vector, followed by a $\tanh$ and another linear projection.

\begin{figure}[t]
  \begin{center}
    \centerline{\includegraphics[width=\columnwidth,scale=0.22,clip=false,trim=0cm 0cm
    0cm 0cm]{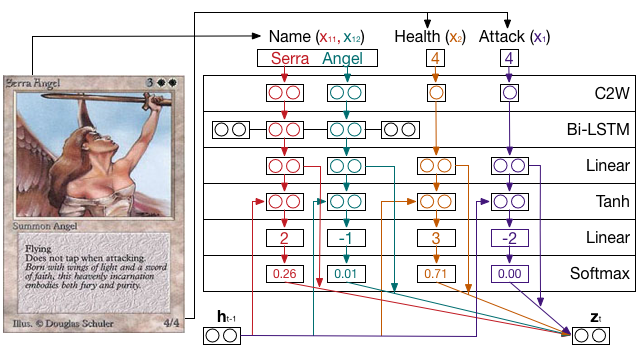}}
    \caption{Illustration of the structured attention mechanism operating on a single time stamp $t$.}
    \label{input_img}
  \end{center}
\end{figure}

\paragraph{Our Approach} We extend the computation of $\mathbf{z}_t$ for cases when $x$ corresponds to multiple fields.
Figure~\ref{input_img} illustrates how the MTG card
``Serra Angel'' is encoded, assuming that there are two singular fields and one
text field. We first encode each token $x_{ki}$ using the C2W model described
in \newcite{wang:2015}, which is a replacement for lookup tables where word
representations are learned at the character level (cf.~\textit{C2W} row). A
context-aware representation is built for words in the text fields using a
bidirectional LSTM (cf.~\textit{Bi-LSTM} row). Computing attention over
multiple input fields is problematic as each input field's vectors have different sizes
and value ranges. Thus, we learn a linear projection mapping each input
token $x_{ki}$ to a vector with a common dimensionality and value range (cf.~\textit{Linear} row).
Denoting this process as $f(x_{ki})$, we extend
Equation~\ref{alignment} as:
\begin{equation}
\label{alignment-mul}
a_{ki} = \underset{x_{ki}\in x}\softmax(v(f(x_{ki}),\mathbf{h}_{t-1}))
\end{equation}
Here a scalar coefficient $a_{ki}$ is computed for each input token
$x_{ki}$ (cf.~``Tanh'', ``Linear'', and ``Softmax'' rows). Thus, the overall input
representation $\mathbf{z}_t$ is computed as:
\begin{equation}
\label{attention-mul}
\mathbf{z}_{t} = \sum_{k = {1..|x|},i = {1..|x_k|}} a_{ij} f(x_{ki})
\end{equation}

\section{Latent Predictor Networks}
\label{sec:lpn}

\begin{figure*}[ht]
\begin{center}
\centerline{\includegraphics[width=2\columnwidth,trim=0cm 0cm
    0cm 0cm]{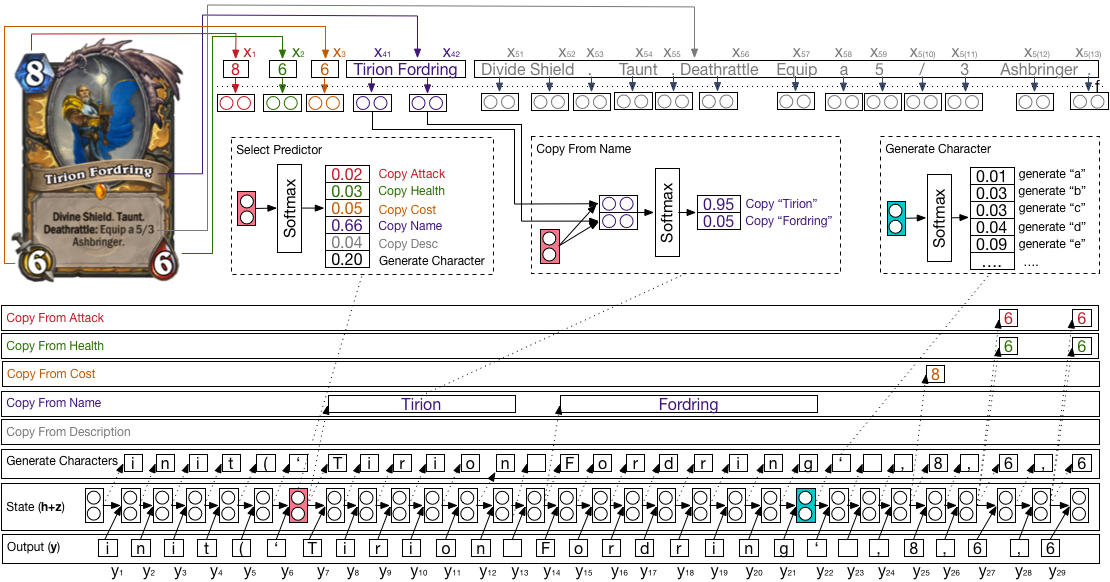}}
\caption{Generation process for the code \texttt{init(`Tirion Fordring',8,6,6)} using LPNs.}
\label{model_img}
\end{center}
\end{figure*}

\paragraph{Background} In order to decode from $x$ to $y$, many words must be copied into the code,
such as the name of the card, the attack and the cost values. If we observe the HS
card in Figure~\ref{hs_img} and the respective code in Figure~\ref{df_code}, we observe that the
name ``Divine Favor'' must be copied into the class name and in the constructor,
along with the cost of the card ``3''.
As explained earlier, this problem is not specific to our task: for instance, in
the dataset of \newcite{oda15ase}, a model must learn to map from \texttt{timeout = int
( timeout )} to ``convert timeout into an integer.'', where the name of the
variable ``timeout'' must be copied into the output sequence. The same issue
exists for proper nouns in machine translation which are typically copied
from one language to the other. Pointer networks~\cite{NIPS2015_5866} address this by
defining a probability distribution over a set of units that can be copied
$c=c_1..c_{|c|}$. The probability of copying a unit $c_i$ is modeled as:
\begin{equation}
\label{copy}
p(c_i) = \underset{c_i\in c}\softmax(v(h(c_i),\mathbf{q}))
\end{equation}
As in the attention model (Equation~\ref{alignment}),
$v$ is a function that computes the affinity between
an embedded copyable unit $h(c_i)$ and an arbitrary vector $\mathbf{q}$.

\paragraph{Our Approach} Combining pointer networks with a character-based softmax is in itself difficult as these generate segments of different granularity and there is no ground truth of which predictor to use at each time stamp.
We now describe Latent Predictor Networks, which model the conditional probability $\log P(y | x)$ over the latent sequence of predictors used to generate $y$.

We assume that our model uses multiple predictors $r\in R$, where each $r$
can generate multiple segments $s_t=y_t..y_{t+|s_t|-1}$ with arbitrary length $|s_t|$ at
time stamp $t$. An example is illustrated in Figure~\ref{model_img}, where we
observe that to generate the code \texttt{init(`Tirion Fordring',8,6,6)}, a
pointer network can be used to generate the sequences $y_7^{13}$=\texttt{Tirion} and
$y_{14}^{22}$=\texttt{Fordring} (cf.~``Copy From Name'' row). These sequences can also be
generated using a character softmax (cf.~``Generate Characters'' row). The same
applies to the generation of the attack, health and cost values as each of these
predictors is an element in $R$. Thus, we define our objective function as a
marginal log likelihood function over a latent variable $\omega$:

\begin{equation}
\label{marginal}
\log P(y \mid x) = \log \sum_{\omega \in \bar{\omega}} P(y, \omega \mid x)
\end{equation}

Formally, $\omega$ is a sequence of pairs $r_t,s_t$, where $r_t\in R$ denotes
the predictor that is used at time\-stamp $t$ and $s_t$ the
generated string. We decompose $P(y,
\omega \mid x)$ as the product of the probabilities of segments $s_t$ and
predictors $r_t$:
\begin{align*}
\nonumber P(y, \omega \mid x) = \prod_{r_t,s_t \in \omega} P(s_t, r_t \mid y_1..y_{t-1}, x) = \\
\prod_{r_t,s_t \in \omega} P(s_t \mid y_1..y_{t-1}, x, r_t) P(r_t \mid y_1..y_{t-1}, x)
\end{align*}
where the generation of each segment is performed in two steps: select the
predictor $r_t$ with probability $P(r_t \mid y_1..y_{t-1}, x)$ and then generate
$s_t$ conditioned on predictor $r_t$ with probability $\log P(s_t \mid y_1..y_{t-1}, x,
r_t)$. The probability of each predictor is computed using a softmax over all
predictors in $R$ conditioned on the previous state $\mathbf{h_{t-1}}$ and the
input representation $\mathbf{z}_t$ (cf.~``Select Predictor'' box). Then, the
probability of generating the segment $s_t$ depends on the predictor type.
We define three types of predictors:

\paragraph{Character Generation} Generate a single character from observed
characters from the training data. Only one character is generated at each
time stamp with probability given by Equation~\ref{char_softmax}.

\paragraph{Copy Singular Field} For singular fields only the
field itself can be copied, for instance, the value of the attack and cost attributes
or the type of card. The size of the generated segment is the
number of characters in the copied field and
the segment is generated with probability $1$.

\paragraph{Copy Text Field} For text fields, we allow each of the words
$x_{ki}$ within the field to be copied. The probability of copying a word is learned with
a pointer network (cf.~``Copy From Name'' box), where $h(c_i)$ is set to
the representation of the word $f(x_{ki})$ and
$\mathbf{q}$ is the concatenation $\mathbf{h}_{t-1}:\mathbf{z}_t$ of the state and input vectors.
This predictor generates a segment with the size of the copied word.

It is important to note that the state vector $\mathbf{h}_{t-1}$ is
generated by building an RNN over the sequence of characters up until the
time stamp $t-1$, i.e.\ the previous context $\mathbf{y}_{t-1}$ is
encoded at the character level. This allows the number of possible states to remain
tractable at training time.

\subsection{Inference}

At training time we use back-propagation to maximize the probability of observed code, according to Equation~\ref{marginal}. Gradient computation must be performed with respect to each computed probability $P(r_t \mid y_1..y_{t-1}, x)$ and $P(s_t \mid y_1..y_{t-1}, x, r_t)$. The derivative $\frac{\partial \log P(y \mid x)}{\partial P(r_t \mid y_1..y_{t-1}, x)}$ yields:
\begin{align*}
\label{derivative}
\frac{\partial \alpha_{t}P(r_t \mid y_1..y_{t-1}, x)\beta_{t,r_t} + \xi_{r_t}}{P(y \mid x) \partial P(r_t \mid y_1..y_{t-1}, x)} &= \frac{\alpha_{t}\beta_{t,r_t}}{\alpha_{|y|+1}}
\end{align*}
Here $\alpha_{t}$ denotes the cumulative probability of all values of $\omega$
up until time stamp $t$ and $\alpha_{|y|+1}$ yields the marginal
probability $P(y \mid x)$. $\beta_{t, r_t} = P(s_t \mid y_1..y_{t-1})\beta_{t+|s_t|-1}$ denotes the cumulative probability
starting from predictor $r_t$ at time stamp $t$, exclusive. This includes the probability of the generated segment $P(s_t \mid y_1..y_{t-1}, x, r_t)$ and the probability of all values of $\omega$ starting from timestamp $t+|s_t|-1$, that is, all possible sequences that generate segment $y$ after segment $s_t$ is produced. For
completeness, $\xi_{r}$ denotes the cumulative probabilities of all $\omega$
that do not include $r_t$. To illustrate this, we refer to Figure~\ref{model_img} and consider the timestamp $t=14$, where the segment $s_{14}=$\texttt{Fordring} is generated. In this case, the cumulative probability $\alpha_{14}$ is the sum of the path that generates the sequence \texttt{init(`Tirion } with characters alone, and the path that generates the word \texttt{Tirion} by copying from the input. $\beta_{21}$ includes the probability of all paths that follow the generation of \texttt{Fordring}, which include $2\times3\times3$ different paths due to the three decision points that follow (e.g. generating 8 using a character softmax vs. copying from the cost). Finally, $\xi_{r}$ refers to the path that generates \texttt{Fordring} character by character.

While the number of possible paths grows exponentially, $\alpha$ and $\beta$ can be computed
efficiently using the forward-backward algorithm for Semi-Markov
models~\cite{NIPS2004_2648}, where we associate $P(r_t \mid y_1..y_{t-1},
x)$ to edges and $P(s_t \mid y_1..y_{t-1}, x, r_t)$ to nodes in the
Markov chain. 

The derivative $\frac{\partial \log P(y \mid x)}{\partial P(s_t \mid
y_1..y_{t-1}, x, r_t)}$ can be computed using the same logic:

\begin{align*}
\frac{\partial \alpha_{t,s_t}P(s_t \mid y_1..y_{t-1}, x, r_t)\beta_{t+|s_t|-1} + \xi_{r_t}}{P(y \mid x) \partial P(s_t \mid y_1..y_{t-1}, x, r_t)} = \\ \frac{\alpha_{t,r_t}\beta_{t+|s_t|-1}}{\alpha_{|y|+1}}
\end{align*}

Once again, we denote $\alpha_{t,r_t}=\alpha_{t}P(r_t \mid y_1..y_{t-1}, x)$ as the cumulative probability of all values of $\omega$ that lead to $s_t$, exclusive.

An intuitive interpretation of the derivatives is that gradient updates will be stronger on probability chains that are more likely to generate the output sequence. For instance, if the model learns a good predictor to copy names, such as \texttt{Fordring}, other predictors that can also generate the same sequences, such as the character softmax will allocate less capacity to the generation of names, and focus on elements that they excel at (e.g. generation of keywords).

\subsection{Decoding}

Decoding is performed using a stack-based decoder with beam search. Each state
$S$ corresponds to a choice of predictor $r_t$ and segment $s_t$ at a given
time stamp $t$. This state is scored as $V(S) = \log P(s_t \mid y_1..y_{t-1}, x,
r_t) + \log P(r_t \mid y_1..y_{t-1}, x) + V(prev(S))$, where $prev(S)$ denotes the
predecessor state of $S$. At each time stamp, the $n$ states with the highest
scores $V$ are expanded, where $n$ is the size of the beam. For each predictor
$r_t$, each output $s_t$ generates a new state. Finally, at each time\-stamp
$t$, all states which produce the same output up to that point are merged by
summing their probabilities.

\section{Code Compression}
\label{sec:compression}

As the attention-based model traverses all input units at each generation step,
generation becomes quite expensive for datasets such as MTG where the average
card code contains 1,080 characters. While this is not the essential contribution in our paper, we propose a simple method to compress the code while maintaining the structure of the code, allowing us to train on datasets with longer code (e.g., MTG).

The idea behind that method is that many keywords in the
programming language (e.g., \texttt{public} and \texttt{return}) as well as frequently
used functions and classes (e.g., \texttt{Card}) can be learned without
character level information. We exploit this by mapping such strings onto
additional symbols $X_i$ (e.g., ~\texttt{public class copy()} $\to$ ``$X_1$
$X_2$ $X_3$()''). Formally, we seek the string $\hat{v}$ among all
strings $V(max)$ up to length $max$ that maximally reduces the size of the
corpus:
\begin{equation}
\hat{v} = \underset{v\in V(max)}\argmax (len(v)-1)C(v)
\end{equation}
where $C(v)$ is the number of occurrences of $v$ in the training corpus and
$len(v)$ its length. $(len(v)-1)C(v)$ can be seen as the number of characters
reduced by replacing $v$ with a non-terminal symbol. To find $q(v)$ efficiently,
we leverage the fact that $C(v) \le C(v')$ if $v$ contains $v'$. It follows that
$(max-1)C(v) \le (max-1)C(v')$, which means that the maximum compression
obtainable for $v$ at size $max$ is always lower than that of $v'$. Thus, if we
can find a $\bar{v}$ such that $(len(\bar{v})-1)C(\bar{v})>(max-1)C(v')$, that is
$\bar{v}$ at the current size achieves a better compression rate than $v'$ at
the maximum length, then it follows that all sequences that contain $v$ can be
discarded as candidates. Based on this idea, our iterative search starts by
obtaining the counts $C(v)$ for all segments of size $s=2$, and computing the
best scoring segment $\bar{v}$. Then, we build a list $L(s)$ of all segments
that achieve a better compression rate than $\bar{v}$ at their maximum size. At
size $s+1$, only segments that contain a element in $L(s-1)$ need to be
considered, making the number of substrings to be tested to be tractable as $s$
increases. The algorithm stops once $s$ reaches $max$ or the newly generated
list $L(s)$ contains no elements.

%
%
%

\begin{table}
\centering
\small
\begin{tabular}{@{}llr@{}}
  \toprule
$X$ & $v$ & size\\
  \midrule
$X_1$ & card)$\Downarrow$\{$\Downarrow$super(card);$\Downarrow$\}$\Downarrow$@Override$\Downarrow$public & 1041\\
$X_2$ & bility & 1002\\
$X_3$ & ;$\Downarrow$this. & 964\\
$X_4$ & (UUID\_ownerId)$\Downarrow$\{$\Downarrow$super(ownerId & 934\\
$X_5$ & public\_ & 907\\
$X_6$ & new\_ & 881\\
$X_7$ & \_copy() & 859\\
$X_8$ & \}")$X_3$expansionSetCode\_=\_" & 837\\
$X_9$ & $X_6$CardType[]\{CardType. & 815\\
$X_{10}$ & ffect & 794\\
  \bottomrule
\end{tabular}
\caption{First 10 compressed units in MTG. We replaced newlines with $\Downarrow$ and spaces with \_.}\label{tab:compress}
\end{table}

Once $\hat{v}$ is obtained, we replace all occurrences of $\hat{v}$ with a new
non-terminal symbol. This process is repeated until a desired average size for the
code is reached. While training is performed on the compressed code, the
decoding will undergo an additional step, where the compressed code is restored
by expanding the all $X_i$. Table~\ref{tab:compress} shows the first
10 replacements from the MTG dataset, reducing its average size from 1080 to 794.


\section{Experiments}
\label{sec:experiments}


\paragraph{Datasets} Tests are performed on the two datasets provided in this
paper, described in Table~\ref{tab:corpus}. Additionally, to test the model's
ability of generalize to other domains, we report results in the Django dataset~\cite{oda15ase},
comprising of 16000 training, 1000 development and 1805 test annotations. Each
data point consists of a line of Python code together with a manually created
natural language description.

\paragraph{Neural Benchmarks} We implement two standard
neural networks, namely a sequence-to-sequence model
~\cite{DBLP:journals/corr/SutskeverVL14} and an attention-based
model~\cite{DBLP:journals/corr/BahdanauCB14}. The former is
adapted to work with multiple input fields by concatenating them,
while the latter uses our proposed attention model. These models
are denoted as ``Sequence'' and ``Attention''.

\paragraph{Machine Translation Baselines} Our problem can also be viewed in the
framework of semantic
parsing~\cite{Wong:2006:LSP:1220835.1220891,Lu:2008:GMP:1613715.1613815,Jones:2012:SPB:2390524.2390593,artzi-lee-zettlemoyer:2015:EMNLP}.
Unfortunately, these approaches define strong assumptions regarding the grammar and
structure of the output, which makes it difficult to generalize for other
domains~\cite{Kwiatkowski:2010:IPC:1870658.1870777}.
However, the work in \newcite{andreas-vlachos-clark:2013:Short} provides evidence
that using machine translation systems without committing to such assumptions
can lead to results competitive with the systems described above.  We follow the
same approach and create a
phrase-based~\cite{Koehn:2007:MOS:1557769.1557821} model and a hierarchical
model (or PCFG)~\cite{Chiang:2007:HPT:1268656.1268659} as benchmarks for the
work presented here.
As these models are optimized to generate words, not characters, we
implement a tokenizer that splits on all punctuation characters, except for
the ``\_'' character. We also facilitate the task by splitting CamelCase words
(e.g., \texttt{class TirionFordring} $\to$ \texttt{class Tirion Fordring}). Otherwise all
class names would not be generated correctly by these methods. We used the models implemented in
Moses to generate these baselines using standard parameters, using IBM Alignment
Model 4 for word alignments~\cite{Och:2003:SCV:778822.778824}, MERT for
tuning~\cite{Sokolov11minimum} and a 4-gram Kneser-Ney Smoothed language
model~\cite{citeulike:13570576}. These models will be denoted as ``Phrase'' and
``Hierarchical'', respectively.


\paragraph{Retrieval Baseline} It was reported in~\cite{quirk:acl15} that a
simple retrieval method that outputs the most similar input for each sample, measured
using Levenshtein Distance, leads to good results.
We implement this baseline by computing the average Levenshtein Distance for each input field.
This baseline is denoted ``Retrieval''.

\paragraph{Evaluation} A typical metric is to compute the accuracy of whether
the generated code exactly matches the reference code. This is informative as
it gives an intuition of how many samples can be used without further
human post-editing. However, it does not provide an illustration on the degree
of closeness to achieving the correct code. Thus, we also test using
BLEU-4~\cite{Papineni:2002:BMA:1073083.1073135} at the token level. There are clearly problems with these metrics. For instance, source code can be correct without matching the reference. The code in Figure~\ref{df_code}, could have also been implemented by calling the \texttt{draw} function in an cycle that exists once both players have the same number of cards in their hands. Some tasks, such as the generation of queries~\cite{zelle:aaai96}, have overcome this problem by executing the query and checking if the result is the same as the annotation. However, we shall leave the study of these methologies for future work, as adapting these methods for our tasks is not trivial. For instance, the correctness cards with conditional (e.g. \texttt{if player has no cards, then draw a card}) or non-deterministc (e.g. \texttt{put a random card in your hand}) effects cannot be simply validated by running the code.

\paragraph{Setup} The multiple input types (Figure~\ref{input_img}) are
hyper-parametrized as follows: The C2W model (cf.~``C2W'' row) used to obtain
continuous vectors for word types uses character embeddings of size 100 and
LSTM states of size 300, and generates vectors of size 300. We also
report on results using word lookup tables of size 300, where we replace
singletons with a special unknown token with probability $0.5$ during training, which is then used for out-of-vocabulary words.
For text fields, the
context (cf.~``Bi-LSTM'' row) is encoded with a Bi-LSTM of size 300 for the
forward and backward states. Finally, a linear layer maps the different input tokens
into a common space with of size 300 (cf.~``Linear'' row). As for the attention model,
we used an hidden layer of size 200 before applying the non-linearity (row
``Tanh''). As for the decoder (Figure~\ref{model_img}), we encode
output characters with size 100 (cf.~``output (y)'' row), and an LSTM state of size 300
and an input representation of size 300 (cf.~``State(h+z)'' row). For each pointer network (e.g., ``Copy From Name'' box),
the intersection
between the input units and the state units are performed with a vector of size 200.
Training is performed using mini-batches of 20 samples using
AdaDelta~\cite{DBLP:journals/corr/abs-1212-5701} and we report results using the
iteration with the highest BLEU score on the validation set (tested at intervals
of 5000 mini-batches). Decoding is performed with a beam of 1000.
As for compression, we performed a grid search over
compressing the code from 0\% to 80\% of the original average length over
intervals of 20\% for the HS and Django datasets. On the MTG dataset, we are forced to
compress the code up to 80\% due to performance issues when training with extremely long sequences.

\subsection{Results}

\begin{table}
   \centering
   \small
   \begin{tabular}{@{}lr@{}rr@{}rr@{}r@{}}
     \toprule
     {\bf } & \multicolumn{2}{c}{\bf MTG} & \multicolumn{2}{c}{\bf HS} &
     \multicolumn{2}{c@{}}{\bf Django}\\
     \cmidrule{2-3}
     \cmidrule{4-5}
     \cmidrule{6-7}
     & {\sc bleu} & \phantom{xx}Acc & {\sc bleu} & \phantom{xx}Acc & {\sc bleu} & \phantom{xx}Acc\\
     \midrule
     Retrieval & 54.9 & 0.0 & 62.5 & 0.0 & 18.6 & 14.7\\
     Phrase & 49.5 & 0.0 & 34.1 & 0.0 & 47.6 & 31.5\\
     Hierarchical & 50.6 & 0.0 & 43.2 & 0.0 & 35.9 & 9.5\\
     \midrule
     Sequence & 33.8 & 0.0 & 28.5 & 0.0 & 44.1 & 33.2\\
     Attention & 50.1 & 0.0 & 43.9 & 0.0 & 58.9 & 38.8\\
     \midrule
     Our System & \textbf{61.4} & \textbf{4.8} & 65.6 & 4.5 & \textbf{77.6} & \textbf{62.3}\\ 
     -- C2W & 60.9 & 4.4 & \textbf{67.1} & 4.5 & 75.9 & 60.9\\
     -- Compress & - & - & 59.7 & \textbf{6.1} & 76.3 & 61.3\\
     -- LPN & 52.4 & 0.0 & 42.0 & 0.0 & 63.3 & 40.8\\
     -- Attention & 39.1 & 0.5 & 49.9 & 3.0 & 48.8 & 34.5\\
      \bottomrule
    \end{tabular}
    \caption{BLEU and Accuracy scores for the proposed task on two in-domain datasets (HS and MTG) and an out-of-domain dataset (Django).}\label{tab:results}
  \end{table}

\paragraph{Baseline Comparison} Results are reported in Table~\ref{tab:results}.
Regarding the retrieval results (cf.~``Retrieval'' row), we observe the best BLEU scores among the
baselines in the card datasets (cf.~``MTG'' and ``HS'' columns).
A key advantage of this method is that retrieving existing entities guarantees that the output is well formed, with no syntactic errors such as producing a non-existent function call or generating incomplete code.
As BLEU penalizes length mismatches, generating code that
matches the length of the reference provides a large boost.
The phrase-based translation model (cf.~``Phrase'' row) performs well in
the Django (cf.~``Django'' column), where mapping from the input to the output is
mostly monotonic, while the hierarchical model (cf.~``Hierarchical'' row) yields
better performance on the card datasets as the concatenation of the input fields needs
to be reordered extensively into the output sequence. Finally, the
sequence-to-sequence model (cf.~``Sequence'' row) yields extremely low results,
mainly due to the lack of capacity needed to memorize whole input and
output sequences, while the attention based model (cf.~``Attention'' row) produces
results on par with phrase-based systems. Finally, we observe that by
including all the proposed components (cf.~``Our System'' row), we obtain
significant improvements over all baselines in the three datasets and is the only one that obtains non-zero accuracies in the card datasets.

\begin{table}
  \centering
  \small
  \begin{tabular}{@{}llrrrrr@{}}
    \toprule
    \multicolumn{2}{@{}l}{Compression} & 0\% & 20\% & 40\% & 60\% & 80\%\\
    \midrule
    \multicolumn{2}{@{}l}{Seconds Per Card} \\
    & Softmax & 2.81 & 2.36 & 1.88 & 1.42 & \textbf{0.94}\\
    & LPN & 3.29 & 2.65 & 2.35 & 1.93 & \textbf{1.41}\\
    \multicolumn{2}{@{}l}{BLEU Scores} \\
    & Softmax & 44.2 & 46.9 & 47.2 & 51.4 & \textbf{52.7}\\
    & LPN & 59.7 & 62.8 & 61.1 & 66.4 & \textbf{67.1}\\
    \bottomrule
  \end{tabular}
  \caption{Results with increasing compression rates with a regular softmax (cf.~``Softmax") and a LPN (cf.~``LPN'').
  Performance values (cf.~``Seconds Per Card'' block) are computed using one CPU.}\label{tab:res_comp}
\end{table}

\paragraph{Component Comparison}
We present ablation results in order to analyze the contribution of each of our modifications.
Removing the C2W model (cf.~``-- C2W'' row) yields a small deterioration, as word lookup tables are more susceptible to sparsity.
The only exception is in the HS dataset, where lookup tables perform better.
We believe that this is because the small size of the training set does not provide enough evidence for the character model to scale to unknown words. 
Surprisingly, running our model compression code (cf.~``-- Compress'' row) actually yields better results.
Table~\ref{tab:res_comp} provides an illustration of the results for different compression rates.
We obtain the best results with an 80\% compression rate (cf.~``BLEU Scores'' block), while maximising the time each card is processed (cf.~``Seconds Per Card'' block).
While the reason for this is uncertain, it is similar to the finding that language models that output characters tend to under-perform those that output words~\cite{DBLP:journals/corr/JozefowiczVSSW16}.
This applies when using the regular optimization process with a character softmax (cf.~``Softmax'' rows), but also when using the LPN (cf.~``LPN'' rows).
We also note that the training speed of LPNs is not significantly lower as marginalization is performed with a dynamic program.
Finally, a significant decrease is observed if we remove the pointer networks (cf.~``-- LPN'' row).
These improvements also generalize to sequence-to-sequence models (cf.~``-- Attention'' row), as the scores are superior to the sequence-to-sequence benchmark (cf.~``Sequence'' row).

\paragraph{Result Analysis}

\begin{figure}[ht]
\begin{center}
\centerline{\includegraphics[width=\columnwidth,trim=0cm 0cm
    0cm 0cm]{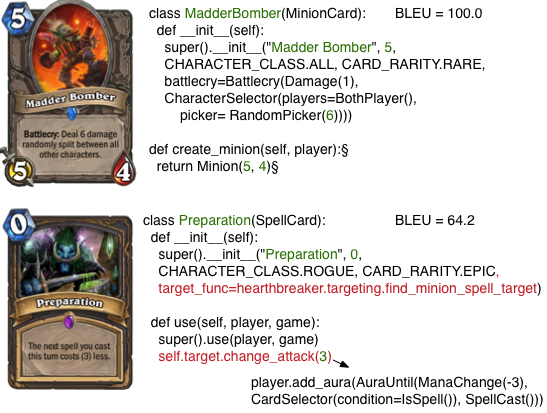}}
\caption{Examples of decoded cards from HS. Copied segments are marked in green and incorrect segments are marked in red.}
\label{exp_img}
\end{center}
\end{figure}

Examples of the code generated for two cards are illustrated in Figure~\ref{exp_img}. We obtain the segments that were copied by the pointer networks by computing the most likely predictor for those segments. We observe from the marked segments that the model effectively copies the attributes that match in the output, including the name of the card that must be collapsed. As expected, the majority of the errors originate from inaccuracies in the generation of the effect of the card. While it is encouraging to observe that a small percentage of the cards are generated correctly, it is worth mentioning that these are the result
of many cards possessing similar effects. The ``Madder Bomber'' card is generated correctly as there is a similar card ``Mad Bomber'' in the training set, which implements the same effect, except that it deals 3 damage instead of 6. Yet, it is a promising
result that the model was able to capture this difference. However, in many cases, effects that radically differ from seen ones tend to be generated incorrectly. In the card ``Preparation'', we observe that while the properties of the card are generated correctly, the effect implements a unrelated one, with the exception of the value 3, which is correctly copied. Yet, interestingly, it still generates a valid effect, which sets a minion's attack to 3. Investigating better methods to accurately generate these effects will be object of further studies.

\section{Related Work}
\label{sec:rel}

While we target widely used programming languages, namely, Java and Python, our
work is related to studies on the generation of any executable code. These include
generating regular expressions~\cite{kushman-barzilay:2013:NAACL-HLT}, and the
code for parsing input documents~\cite{lei-EtAl:2013:ACL2013}. Much research
has also been invested in generating formal languages, such as database
queries~\cite{zelle:aaai96,BerantCFL13}, agent specific language~\cite{kate:aaai05} or smart phone instructions~\cite{Le:2013:SSS:2462456.2464443}. Finally, mapping natural language into a sequence of actions for the generation of executable code~\cite{Branavan:2009:RLM:1687878.1687892}.
Finally, a considerable effort in this task has focused on
semantic
parsing~\cite{Wong:2006:LSP:1220835.1220891,Jones:2012:SPB:2390524.2390593,lei-EtAl:2013:ACL2013,artzi-lee-zettlemoyer:2015:EMNLP,quirk:acl15}.
Recently proposed models focus on Combinatory Categorical
Grammars~\cite{kushman-barzilay:2013:NAACL-HLT,artzi-lee-zettlemoyer:2015:EMNLP},
Bayesian Tree
Transducers~\cite{Jones:2012:SPB:2390524.2390593,lei-EtAl:2013:ACL2013} and
Probabilistic Context Free Grammars~\cite{andreas-vlachos-clark:2013:Short}. The
work in natural language
programming~\cite{vadas-curran:2005:ALTA20052,conf/aaai/ManshadiGA13}, where
users write lines of code from natural language, is also related to our work.
Finally, the reverse mapping from code into natural language is explored in~\cite{oda15ase}.

Character-based sequence-to-sequence models have previously been used to generate code from natural language in~\cite{DBLP:journals/corr/MouMLZJ15}.
Inspired by these works, LPNs provide a richer framework by employing attention models~\cite{DBLP:journals/corr/BahdanauCB14}, pointer networks~\cite{NIPS2015_5866} and character-based embeddings~\cite{wang:2015}.
Our formulation can also be seen as a generalization of ~\newcite{2016arXiv160203001A}, who implement a special case where two predictors have the same granularity (a sub-token softmax and a pointer network).
Finally, HMMs have been employed in neural models to marginalize over label sequences in~\cite{Collobert:2011:NLP:1953048.2078186,2016arXiv160301360L} by modeling transitions between labels.

\section{Conclusion}
\label{sec:conclusion}

We introduced a neural network architecture named \emph{Latent Prediction Network}, which allows efficient marginalization over multiple
predictors. Under this architecture, we propose a generative model for code generation that combines a character level softmax to generate language-specific tokens and multiple pointer networks to copy keywords from the input. Along with other extensions, namely structured attention and code compression, our model is applied on on both existing datasets and also on a newly created one with implementations of TCG game cards. Our experiments show that our model out-performs multiple benchmarks, which demonstrate the importance of combining different types of predictors.



\bibliography{acl2016}
\bibliographystyle{acl2016}

\appendix

\end{document}